\documentclass[letterpaper, 10 pt, conference]{ieeeconf}  

\IEEEoverridecommandlockouts                              

\usepackage{amsmath} 
\usepackage{amssymb}  

\usepackage{amsmath} 
\usepackage{float}
\usepackage{color}
\usepackage[utf8]{inputenc}    
\usepackage[T1]{fontenc}       
\usepackage{cite}
\makeatletter
\let\NAT@parse\undefined
\makeatother
\usepackage[colorlinks = true,
            linkcolor = black,
            urlcolor  = blue,
            citecolor = black,
            anchorcolor = black]{hyperref}
\pdfoutput=1
\hypersetup{bookmarksdepth=1}
\usepackage[font={small}]{caption}

\usepackage{subcaption}
\usepackage{float}
\usepackage{fixmath}
\usepackage{xcolor}
\usepackage{tikz}
\usepackage{pgfplots}
\pgfplotsset{compat=newest}
\usepgfplotslibrary{groupplots}
\let\labelindent\relax
\usepackage{enumitem}

\title{\LARGE \bf
Does Unpredictability Influence Driving Behavior?
}

\author{Sepehr Samavi$^{1}$, Florian Shkurti$^{2}$, and Angela P. Schoellig$^{3}$
\thanks{\vspace{-0.0cm}%
Computing resources used in preparing this research were provided, in part, by the Province of Ontario, the Government of Canada through CIFAR, and companies sponsoring the Vector Institute for Artificial Intelligence.}
\thanks{$^{1}$University of Toronto Institute for Aerospace Studies (UTIAS) and the Vector Institute. {\tt\footnotesize sepehr@robotics.utias.utoronto.ca}.}%
\thanks{$^{2}$Department of Computer Science at the University of Toronto
        {\tt\footnotesize florian@cs.toronto.edu}.}%
\thanks{$^{3}$Technical University of Munich, UTIAS and the Vector Institute. {\tt\footnotesize angela.schoellig@tum.de}.}%
}

\begin{document}

\maketitle
\thispagestyle{empty}
\pagestyle{empty}

\begin{abstract}
In this paper we investigate the effect of the unpredictability of surrounding cars on an ego-car performing a driving maneuver. We use Maximum Entropy Inverse Reinforcement Learning to model reward functions for an ego-car conducting a lane change in a highway setting. We define a new feature based on the unpredictability of surrounding cars and use it in the reward function. We learn two reward functions from human data: a baseline and one that incorporates our defined unpredictability feature, then compare their performance with a quantitative and qualitative evaluation. Our evaluation demonstrates that incorporating the unpredictability feature leads to a better fit of human-generated test data. These results encourage further investigation of the effect of unpredictability on driving behavior.
\end{abstract}

\vspace{-0.1cm}
\section{Introduction}
An increasing number of autonomous vehicles (AVs) have started to be deployed in mixed traffic settings, where the AV shares the same road with human-driven vehicles. %
To be successful in these settings, the AV needs to understand and predict the continuous chain of interactions between drivers.
Specifically, the AV needs to be able to predict the behavior of other vehicles to seamlessly interact and arrive at its goal. %

In a typical AV autonomy stack, a prediction model generates information about the future of the surrounding environment (e.g. location of adjacent vehicles, pedestrians, road geometry) and supplies that information into a planning module to produce collision-free trajectories \cite{Falcone2007, Schildbach2015, Liniger2015, Burnett2020}. Therefore, the accuracy of the prediction model is critical for safe AV operation. To this end, there has been much progress in enabling accurate predictions \cite{Alahi2016a, Deo2018, Ivanovic2019}. However, any model comes with uncertainty, and thus a number of recent methods have focused on quantifying the uncertainty and/or confidence of prediction models \cite{fridovich2020confidence, Sun2021OnCE}. The goal of this uncertainty quantification is to avoid dangerous planning caused by inaccurate predictions by switching to another prediction model \cite{Sun2021OnCE}, or switching to a more conservative planner \cite{katyal2020intent}. While most methods investigate how to incorporate uncertainty into a robot's autonomy stack, %
we explore whether predictive uncertainty can provide insight on the trajectories taken by human drivers in scenarios where an adjacent driver is behaving erratically, and thus unpredictably.
\begin{figure}[t]
    \includegraphics[width=\columnwidth]{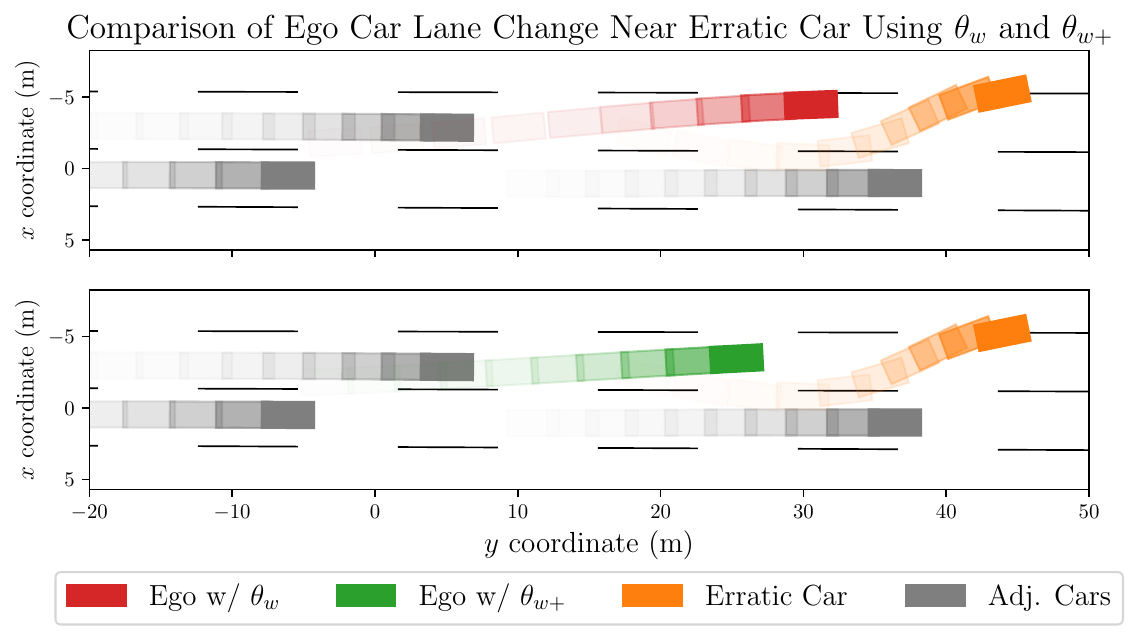}
    \centering
    \captionsetup{skip=0.1pt}
    \caption{We compare a snapshot (at $t=3s$) of ego car lane change trajectories generated using the baseline (red) and the unpredictability-aware (green) models, for identical adjacent cars (gray and orange) and lanes. The history of each vehicle is illustrated with decreasing opacity. The erratic adjacent vehicle is perturbed to zigzag. We observe that the trajectory from the unpredictability-aware model (green) delays changing into the target lane and maintains a higher distance from the erratic car (orange) than the baseline (red). Video at \href{https://tiny.cc/unpredi}{\texttt{tiny.cc/unpredi}}.}
    \label{fig:high_unpred_artificial}
    \vspace{-0.75cm}
\end{figure}

In this paper we propose {\it unpredictability} as a feature for a trajectory planning reward function. For an ego-vehicle performing a trajectory, we measure the unpredictability of each interacting adjacent car by using %
the performance metrics of an off-the-shelf vehicle trajectory prediction model \cite{Deo2018}. %
We assume that if the off-the-shelf model performs poorly on a particular adjacent car at a particular time, then that car is behaving unpredictably and an ego-car should keep its distance. Our intuition is captured in Fig.~\ref{fig:high_unpred_artificial}, where we illustrate a snapshot of ego cars (red and green) performing a simulated lane change in an identical scenario involving an erratic adjacent car (orange). In the top figure, we can see that in a trajectory generated without accounting for unpredictability, the ego-car (red) maintains a lower distance from the erratic car compared to the one in the bottom figure where unpredictability is accounted for (green).

The main contribution of this paper is defining a reward feature incorporating unpredictability and analyzing human lane change behavior using an Inverse Reinforcement Learning (IRL) approach. %
Using IRL, we learn two reward functions for generating lane change trajectories: an unpredictability-aware reward that includes the defined feature and a baseline reward that does not. We demonstrate that, compared to the baseline, the unpredictability-aware reward generates trajectories that are more similar to those in a human-generated dataset. %
We also analyze the qualitative similarities between trajectories generated by the two rewards and human trajectories in test scenarios with variable unpredictable behavior by adjacent cars. This analysis provides insights into the effect of the unpredictability feature.
\section{Related Work}
\label{sec:related-work}
Modelling human driving behavior is explored from different perspectives in AV research.
\subsubsection*{Prediction and Forecasting}
In the classical modular AV design, the output of a prediction model is used in motion planning \cite{Falcone2007, Schildbach2015, Liniger2015}.
In the field of prediction models for AVs, many works use camera fed directly into Deep Neural Network (DNN) architectures such as Long Short-Term memory (LSTM) recurrent neural network architectures \cite{Lotter2016, Villegas2017, Byeon2018} or Variational Auto Encoders (VAEs) \cite{Babaeizadeh2017, Rhinehart2019, Xue2016} to forecast the evolution of a scene.
Other proposals use similar network architectures with a variety of sensors, such as cameras or LIDAR to represent the surrounding world as a Dynamic Occupancy Grid Map (DOGMa) in 2D \cite{Hoermann2018, Mohajerin2019, Itkina2019} or 3D \cite{LuoWenjieandYangBinandUrtasun2018}.
A third group of prediction models focus on predicting time-series trajectories (future positions and velocity) \cite{Alahi2016a, Deo2018, salzmann2020trajectron++}. In order to produce more accurate predictions, some works explicitly account for interactions between the agents \cite{Alahi2016a, Deo2018}, and even account for the effect of an ego AV's plans on the predictions of future behavior by other agents \cite{salzmann2020trajectron++}.
We propose tracking the predictive accuracy of the model proposed in \cite{Deo2018}, which predicts trajectories of vehicles driving on a highway, as an unpredictability measure that we incorporate in a reward feature we describe in Section~\ref{sec:unpred-metric}.
\subsubsection*{Imitation Learning Approaches}
Other methods focus on imitating human driving styles by learning AV planning models that imitate human behavior from data. A common approach is to learn a parametric reward function represented a weighted linear combination of nonlinear features \cite{Levine2012, Kuderer2015} or a DNN \cite{Wulfmeier2015} using a Maximum Entropy Inverse Reinforcement Learning (MaxEnt IRL) learning approach \cite{Ziebart2008}, then use the reward function in a separate planner to optimize trajectories that imitate human behavior. In contrast, end-to-end methods forgo explicitly learning a reward function and learn policies directly from human data \cite{Hawke2020,Huang2020}.

\subsubsection*{Feature-design Approaches}
In contrast to the proposals presented above that focus on \textit{how} to generate driving behavior, our work focuses on investigating \textit{what} features can be used to generate driving behavior. Among methods that focus on feature design, the authors in \cite{Buckman2020} propose maximizing the visibility of the ego-car in highway and blind intersection scenarios. In a driving simulation, they demonstrate that a visibility-aware AV slows down before entering another vehicle's blind spots.
In \cite{Sun2018}, the authors formalize a notion of being courteous to other interacting drivers. They hypothesize that human drivers are in fact also courteous. The authors test their hypothesis using an IRL analysis, in the same way as this paper intends to do. Rather than courtesy, we define the unpredictability of surrounding traffic as a feature to incorporate in trajectory planning.

\subsubsection*{Uncertainty and Confidence}
The concept of unpredictability has also been studied in the context of estimating model uncertainty and confidence for prediction models in AVs and other robotics applications. %
Uncertainty estimation methods include using Bayesian techniques to track a parameter that governs the variance of predictions \cite{fridovich2020confidence}, comparing the output from an ensemble models \cite{Sun2021OnCE}, and tracking the difference between predictions and actual outcomes \cite{sripathy2021dynamicallyswitching, katyal2020intent}. The motivation behind these methods is to use the estimate to enable downstream planning and control of the AV to avoid dangerous actions caused by low-confidence predictions.

In this paper, however, we investigate whether incorporating an estimate of unpredictability into a trajectory planner could have the additional benefit of generating plans that are more similar to human ones by generating more conservative behavior when faced with an unpredictable adjacent vehicle.
The authors in \cite{Hayashi2019} have shown that it is possible to learn a driving policy by maximizing only the predictive accuracy of the behavior of adjacent vehicles as the reward in a Reinforcement Learning (RL) approach. However, their model is exclusively trained and tested in a simulator, and does not use any human data. Therefore, the notion of predictability in their proposal is limited to the driver models that coded in the simulator. We focus on real human trajectory data and compare trajectories generated by using a baseline reward and an unpredictability-aware reward with trajectories from a dataset of human lane changes.

\section{Problem Formulation} \label{sec:prob-form}
\subsection{Lane Change Planning Problem} \label{sec:lc-planning-problem}
We formulate a lane change maneuver as trajectory optimization. Concretely, we model the maneuver as a deterministic discrete-time optimal control task over a finite horizon, $K$, and continuous state and action spaces.
Starting from an initial state, $\mathbf{x}_0$, that is known a priori, the optimization problem can be written as,
\begin{equation} \label{eq:fwd-traj-opt}
	\begin{aligned}
		\max_{\mathbf{x}, \mathbf{u}} & \sum_{k=0}^{K-1} \mathcal{R}(\mathbf{x}_k, \mathbf{u}_k) \\
		\text{ subject to } & \mathbf{x}_{k+1} = {f}(\mathbf{x}_k,\mathbf{u}_k), \text{  } k = 0 \dots K-1
	\end{aligned}
\end{equation}
where $(\mathbf{x}, \mathbf{u})=([\mathbf{x}_1^\top, \dots, \mathbf{x}_K^\top]^\top, [\mathbf{u_0}^\top, \dots, \mathbf{u}_{K-1}^\top]^\top)$ are the sequences of states and actions in lifted notation, $\mathcal{R}$ is the reward function, and ${f}$ is the ego-vehicle dynamics. %

\subsubsection*{System and Dynamics} \label{sec:lc-sys}
We model the ego-vehicle with a kinematic unicycle model. %
The state is $\mathbf{x}_k = \begin{bmatrix} x_k,	y_k, \psi_k	\end{bmatrix}^\top \in \mathbb{R}^3$, where $(x_k,y_k)$ is the two-dimensional position of the ego-vehicle on a 2D map and $\psi_k$ is its heading at time step $k$. The control action is defined as $\mathbf{u}_k = \begin{bmatrix} v_k, \omega_k	\end{bmatrix}^\top \in \mathbb{R}^2$, where $v_k$ is the longitudinal velocity input and $\omega_k$ is the steering input of the car. We discretize time using the forward Euler method with a sampling rate of $\delta t = 0.1 s$ to obtain the dynamics model,
\begin{equation}
	\label{eq:dynamics}
	f(\mathbf{x}_k,\mathbf{u}_k) = \mathbf{x}_k + \delta t \begin{bmatrix}
	v_k cos(\psi_k), %
	v_k sin(\psi_k), %
	\omega_k %
\end{bmatrix}^\top.
\end{equation}

\subsubsection*{Reward Function} \label{sec:lc-reward}
The reward function is formulated as a linear combination of nonlinear reward features,
\begin{equation} \label{eq:reward-struct}
	\mathcal{R}_k(\mathbf{x}_k, \mathbf{u}_k) = \mathbold{\theta}^\top \mathbold{\phi}(\mathbf{x}_k, \mathbf{u}_k),
\end{equation}
where $\mathbold{\phi} := [\phi_1, \dots, \phi_p]^\top$ are a set of $p$ features and each feature is defined as a nonlinear function that incorporates elements of the ego-vehicle's surroundings that are relevant to performing a lane change maneuver.
The parameters $\mathbold{\theta} := [\theta_1, \dots, \theta_p]^{\top} \in \mathbb{R}^{p}_{\geq 0}$ provide the weight of each feature in $\mathbold{\phi}$ relative to the others.%

We reviewed features commonly used for trajectory generation in highway driving. The authors in \cite{Naumann2020} present a detailed survey of features.
The features that compose our baseline reward function are enumerated below,
\begin{enumerate}[leftmargin=0.5cm]
	\item \emph{Lateral deviation from target lane} \cite{Sun2018} is represented as,
	\begin{equation} \label{eq:phi_d}
	   \phi_d := -\exp \left ( \frac{d_k}{w} \right ),
  \end{equation}
	where $d_k$ is the distance from the position of the car, $\mathbf{x}_k$, to the centerline of the target lane. We parametrize the target-lane as a line and $w$ is the average distance between the centerlines of the current-lane and target-lane in the vicinity of the lane change maneuver.
	\item \emph{Deviation from the mean speed of traffic} is represented as, %
	\begin{equation} \label{eq:phi_v}
		\phi_v := -(v_k - v_d)^2,
	\end{equation}
	where $v_d$ is the mean speed of the 4 closest adjacent vehicles over the entire trajectory.
	\item \emph{High angular speed} is represented as,
	\begin{equation}
		\phi_a := -\omega_k^2.
	\end{equation}
	\item \emph{Time To Collision (TTC) with preceding vehicles} is represented as,
	\begin{equation} \label{eq:prec_feature}
		\phi_{p}\! :=\! -\sum_{i}\! h_1(\alpha_{i,k}) \exp \left (-\frac{1}{t_{p}^2} \frac{||\mathbf{x}_k - \mathbf{x}_{i,k}||^2}{v_k^2} \right),
	\end{equation}
	where $i$ is the index of the two preceding vehicles, and $\alpha_{i,k}$ is the angle between the heading of the ego-car and the vector $\mathbf{x}_{i,k} - \mathbf{x}_k$. The function $h_1(\alpha_{i,k}) = \exp\left (-c \alpha_{i,k} \right) $ when $-\frac{\pi}{2} \leq \alpha_{i,k} \leq \frac{\pi}{2}$, and $0$ otherwise, restricts penalizing TTC to preceding vehicles that are \textit{in front} of the ego-vehicle. The constants $c$ and $t_{p}$ are hyperparameters.
	\item \emph{Time To Collision (TTC) with following vehicle} is represented as,
	\begin{equation} \label{eq:foll_feature}
		\phi_{f}\! :=\! - h_2(\mathbf{x}_k) \exp \left (-\frac{1}{t_{f}^2} \frac{||\mathbf{x}_{f,k} - \mathbf{x}_k||^2}{v_{f,k}^2} \right),
	\end{equation}
	where $\cdot_f$ is the index of the vehicle that follows behind the ego-car in the target lane, and $v_{f,k}$ is its speed. The function $h_2(\mathbf{x}_k) = \frac{d_k^2}{w^2}$ (with $d_k$ as in \eqref{eq:phi_d}) restricts penalizing the TTC to the duration when the ego-vehicle is \textit{crossing into} the target lane (i.e. $h_2(\cdot) > 0$), but not after being safely \textit{in} front of the following vehicle.
\end{enumerate}

\subsection{Unpredictability Metric and Reward Feature} \label{sec:unpred-metric}
Most AVs that operate in human environments, use a prediction model of the other agents' behavior \cite{Schwarting2018}. Our approach to measuring predictability – which we call \textit{unpredictability} – uses the tracked performance of such prediction models throughout a maneuver. We propose to take an off-the-shelf prediction model that has been learned from human data \cite{Deo2018}, then to use the performance of that model as a heuristic for measuring predictability.

We calculate the mean Euclidean predictive error of the prediction with respect to the ground truth and use this value as the unpredictability metric.
For adjacent vehicle $i$ at time $k$, we evaluate the predictive error of the prediction made by the model at time $k-t_n$. Then take the average of the error from time $k-t_n$ to $k$ as the unpredictability measure.
Concretely, for adjacent car $i$ at time step $k$ the mean Euclidean predictive error is,
\begin{equation}
	z_{i,k}\!\! := \!\frac{1}{t_n} \!\! \sum_{j = k-t_n+1}^k \!\!\! \sqrt{(x_{i,j} - \hat{x}^{(k-t_n)}_{i,j})^2 + (y_{i,j} - \hat{y}^{(k-t_n)}_{i,j})^2},
	\label{eq:unpred-metric}
\end{equation}
where $t_n$ is the number of time steps we look to the past, $0.2s$ in our implementation, $(\hat{x}^{(k-t_n)}_{i,j}, \hat{y}^{(k-t_n)}_{i,j})$ is the prediction made at time $k-t_n$ by the model \cite{Deo2018} of where car $i$ would be at time $j$, and $(x_{i,j}, y_{i,j})$ is the observed position of car $i$ at time $j$.

In order to incorporate the measure of unpredictability into our baseline lane change reward function, we augment the reward function with two additional features analogous to $\phi_p$ and $\phi_f$,
\begin{enumerate}[leftmargin=0.5cm]
	\setcounter{enumi}{6}
	\item \emph{Unpredictability-weighted TTC with preceding vehicles}.
	\begin{equation} \label{eq:prec_z_feature}
		\phi_{pz}\! :=\! - \!\! \sum_{i}\! h_1(\alpha_{i,k}) \exp\! \left(\!-\frac{1}{t_{p}^2} \frac{||\mathbf{x}_k - \mathbf{x}_{i,k}||^2\!-\!c_{p}z_{i,k}^2}{v_k^2}\right),
	\end{equation}

	\item \emph{Unpredictability-weighted TTC with following vehicle}. Analogously, we define
	\begin{equation} \label{eq:foll_z_feature}
		\phi_{fz}\! :=\! - h_2(\mathbf{x}_k) \exp\!\left(\!-\frac{1}{t_{f}^2} \frac{||\mathbf{x}_{f,k}\!\! - \!\!\mathbf{x}_k||^2 - c_{f}z_{f,k}^2}{v_{f,k}^2}\right),
	\end{equation}
\end{enumerate}
The intuition behind our formulation is that, if an adjacent vehicle is behaving unpredictably, we expect the driver of the ego-car to weigh a collision with them more highly. As such, we use the unpredictability error \eqref{eq:unpred-metric} multiplied by a constant ($c_p$ and $c_f$, respectively) as an additional dynamic weight on the baseline features.
\vspace{-0.1cm}
\section{Experimental Methodology}
\vspace{-0.1cm}
\label{sec:methodology}
The primary focus of this paper is to investigate the effect of incorporating unpredictability on generated lane change trajectories. %
We hypothesize that incorporating the unpredictability feature in a lane change planner produces trajectories that better fit to dataset of human lane changes. We use an IRL approach to demonstrate this effect. %
\subsubsection*{Independent Variable}
We learn two reward functions using the IRL algorithm that we will describe in Section~\ref{sec:irl-theory}, on the same human lane change training data, $\{(\mathbf{x}_{e}^{(i)}, \mathbf{u}_{e}^{(i)})\}^{N}_{i=0}$. We learn the vector $\mathbold{\theta}_{w} \in \mathbb{R}^5_{\geq 0}$ that
parametrizes the baseline reward function described in Section~\ref{sec:lc-reward}, and the vector $\mathbold{\theta}_{w+} \in \mathbb{R}^7_{\geq 0}$ that parametrizes the reward function that includes the additional unpredictability feature, described in Section~\ref{sec:unpred-metric}. %
\subsubsection*{Dependent Variable}
We generate two sets of trajectories that are optimal under the respective learned reward functions, $\{(\mathbf{x}_{w}^{(i)}, \mathbf{u}_{w}^{(i)})\}^{{N}_{\text{test}}}_{i=0}$ and $\{(\mathbf{x}_{w+}^{(i)}, \mathbf{u}_{w+}^{(i)})\}^{{N}_{\text{test}}}_{i=0}$, in identical test scenarios. %
Then, for each test scenario, we compare the two generated trajectories from the learned models with the human expert trajectory from that scenario by calculating Mean Euclidean Error (MEE). For the test dataset, we calculate the Average MEE, $\overline{MEE}_{w} = \frac{1}{{N_{\text{test}}}} \sum_{i=1}^{N_{\text{test}}} \sum_{k=1}^{K} ||\mathbf{x}_{w}^{(i)} - \mathbf{x}_{e}^{(i)}||$ for the baseline, and $\overline{MEE}_{w+}$ for the unpredictability-aware model. %
\vspace{-0.1cm}
\subsection{Lane Change Dataset Preprocessing}
We use the human driving datasets, NGSIM US-101 \cite{Colyar2006} and I-80 \cite{Colyar2006a} from the USA and the highD \cite{Krajewski2018} from Germany, which contain position, time, and lane label information for each vehicle. The I-80 and US-101 datasets are each split into three sections  with increasing traffic congestion, $t0$, $t1$, and $t2$.

\subsubsection*{Extracting Lane Changes}
We denoise the dataset using a symmetric exponential moving filter to smooth the paths \cite{Thiemann2008}. For each path in each dataset, we use finite differences to calculate heading and linear and angular velocity. %
We search the dataset for changes in the lane label of a vehicle, $t_{lc}$, and extract the trajectory of the ego-car conducting the lane change on the interval $t \in [t_{lc}-2.0, t_{lc}+5.0]$, which is the typical duration of lane changes on the datasets \cite{Thiemann2008}. We also extract the paths of four adjacent vehicles: two vehicles preceding the ego-car before and after crossing into the target lane, and two vehicles following the ego-vehicle before and after crossing into the target lane.
We characterize the \textit{current-lane} and the \textit{target-lane} by performing linear least squares regression on the paths of all vehicles assigned to each lane across the entire dataset in the vicinity of the lane change.
We normalize all position values %
such that every trajectory begins at $\mathbf{x}_0 = [0,0,\psi_0]^\top$. %
\subsubsection*{Training, Validation, Testing Split}
We perform training, validation, and testing set splitting for the lane change IRL models such that the sets match the split used to train, validate and test the off-the-shelf prediction model \cite{Deo2018} that we use for the unpredictability feature described in Section~\ref{sec:unpred-metric}.
Only the lane changes used in the training set for the off-the-shelf prediction model are included in our lane change IRL training set.
\vspace{-0.1cm}
\subsection{Learning Lane Change Reward Function from Humans} \label{sec:irl-theory}
\begin{figure*}[th]
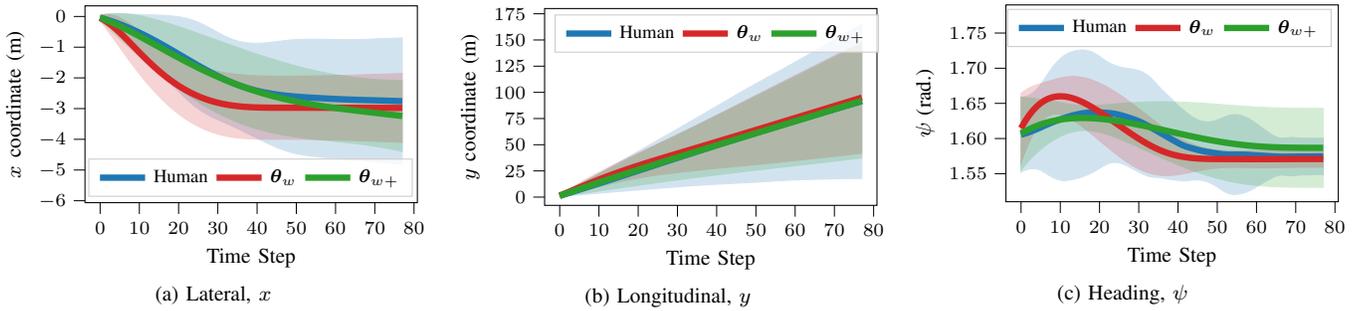

	\centering
	\footnotesize
	\begin{subfigure}[]{0.32\textwidth}
		\centering
	    \footnotesize
		\input{figs/split_up_trajs_NGSIM_smooth5_unpred7_cspNomEp40_smooth5_tpc2_Test_usa101_t1_all_avg_x.tex}
	    \vspace{-0.4cm}
		\caption{Lateral, $x$}
	    \label{fig:real_good_test_avg_x}
	\end{subfigure}
	\hfill
	\begin{subfigure}[]{0.32\textwidth}
		\centering
	    \footnotesize
		\input{figs/split_up_trajs_NGSIM_smooth5_unpred7_cspNomEp40_smooth5_tpc2_Test_usa101_t1_all_avg_y.tex}
		\vspace{-0.4cm}
		\caption{Longitudinal, $y$}
	    \label{fig:real_good_test_avg_y}
	\end{subfigure}
	\hfill
	\begin{subfigure}[]{0.32\textwidth}
		\centering
	    \footnotesize
		\input{figs/split_up_trajs_NGSIM_smooth5_unpred7_cspNomEp40_smooth5_tpc2_Test_usa101_t1_all_avg_psi.tex}
	    \vspace{-0.4cm}
		\caption{Heading, $\psi$}
	    \label{fig:real_good_test_avg_psi}
	\end{subfigure}
	\caption{State values compared to time. We compare the human ego-car lane change trajectory (blue) with trajectories generated by optimizing rewards with baseline reward, $\mathbold{\theta}_{w}$ (red), and unpredictability-aware reward, $\mathbold{\theta}_{w+}$ (green) on the US-101 $t1$ test dataset. We plot the average (solid lines) and 3-$\sigma$ bounds (translucent fill) of the  values. In the lateral direction (a) and heading (c), we observe more similarity between the $\mathbold{\theta}_{w+}$ model (green) and the human data (blue) compared to the $\mathbold{\theta}_{w}$ model (red).}
	\label{fig:test_set_avg}
	\vspace{-0.4cm}
\end{figure*}

\subsubsection*{IRL Algorithm}
Given training set of $N$ human lane change trajectories, $\{(\mathbf{x}_{e}^{(i)}, \mathbf{u}_{e}^{(i)})\}^{N}_{i=0}$, we obtain a maximum likelihood estimate of reward parameters, $\mathbold{\theta_{(\cdot)}}$ \cite{Levine2012}.

We define a probability distribution over lane change trajectories as,
\vspace{-0.35cm}
\begin{equation} \label{eq:distribution}
	\mathrm{p}(\mathbf{u}|\mathbf{x}_0) = \frac{ \exp \left( \sum_{k=0}^{K-1} \mathcal{R}_k(\mathbf{x}_k, \mathbf{u}_k) \right)}{\int \exp \left( \sum_{k=0}^{K-1} \mathcal{R}_k(\tilde{\mathbf{x}}_k, \tilde{\mathbf{u}}_k) \right)  d\tilde{\mathbf{u}}}.
\end{equation}
\vspace{-0.05cm}
The denominator is the sum of every dynamically feasible trajectory and intractable to calculate due to the continuous state and action spaces. Since the dynamics are deterministic, we can find $\mathbf{x}$ given $\mathbf{u}$ and $\mathbf{x}_0$, thus the distribution can be written as a function of $\mathbf{u}$. Following \cite{Levine2012} we use the Laplace approximation to find an approximate solution to the denominator. We arrive at a closed-form estimate of the probability,
\vspace{-0.25cm}
\begin{equation} \label{eq:prob-cioc} %
\begin{aligned}
	\mathrm{p}(\mathbf{u} | \mathbf{x}_0) &\approx e^{\frac{1}{2} \mathbf{g}^\top \mathbf{H}^{-1} \mathbf{g}} |-\mathbf{H}|^{\frac{1}{2}} (2 \pi)^{-\frac{d_{\mathbf{u}}}{2}},
\end{aligned}
\end{equation}
where $\mathbf{g}:=\frac{\partial \mathcal{R}}{\partial \mathbf{u}}$ is the gradient of the reward with respect to the actions, $\mathbf{H}:=\frac{\partial^2 \mathcal{R}}{\partial \mathbf{u}^2}$ is the Hessian with respect to the actions, $|\cdot|$ denotes the determinant, and $d_{\mathbf{u}}$ denotes the dimension of the lifted vector $\mathbf{u} \in \mathbb{R}^{Km}$. We finally obtain the approximate log likelihood,
\begin{equation} \label{eq:likelihood-cioc}
	\mathcal{L} = \frac{1}{2} \mathbf{g}^\top \mathbf{H}^{-1} \mathbf{g}	+ \frac{1}{2	}\log|-\mathbf{H}| - \frac{d_{\mathbf{u}}}{2}\log(2 \pi).
\end{equation}%
Assuming the $N$ trajectories in a dataset are independent and identically distributed (i.i.d), we can sum the likelihood of each trajectory to get the likelihood over the entire dataset then optimize the reward parameters $\mathbold{\theta}$ that maximize the likelihood of the dataset under the model.
\subsubsection*{IRL Implementation Details}
We use the Sequential Least-Squares Quadratic Program (SLSQP) algorithm \cite{Kraft1994} from the NLOpt package (v2.6.2) \cite{JohnsonNLopt}. We calculate the $\mathbf{g}$ and $\mathbf{H}$ in addition to the gradient of \eqref{eq:likelihood-cioc} automatically using JAX \cite{jax2018github}. We perform a sweep over hyperparameters defined in Section~\ref{sec:prob-form}, and use the value that performs best on the training set.
For numerical stability, we use min-max normalization on the features, exploit the identity $\log|\mathbf{-H}|=2\mathrm{tr}(\mathbf{L})$ where $\mathbf{L}$ is the Cholesky factor, %
and regularize $\mathbf{H}$ to be negative definite  \cite{Levine2012}.

\section{Results and Discussion} \label{sec:results}

\begin{table}[t]
	\footnotesize
	\centering
    \captionsetup{skip=0pt}
	\caption{\footnotesize Model Performance on Training Datasets}
	\label{tab:results-combined-train}
	\begin{tabular}{|l|l|l|l|l|l|}
		\hline
        Highway  & \# Traj. & $\overline{MEE}_{w} \downarrow$ & $\overline{MEE}_{w+} \downarrow$ & \% Imp $\uparrow$    \\ \hline
		i80 $t0$	& 188  & $3.77 \pm 2.53$  & $3.73 \pm 2.58$ & $1.17 \%$ \\ \hline
		i80 $t1$ 	& 152  & $3.87 \pm 2.34$  & $3.73 \pm 2.32$ & $3.68 \%$ \\ \hline
		i80 $t2$	& 180  & $3.50 \pm 2.37$  & $3.46 \pm 2.32$ & $1.00 \%$ \\ \hline
		us101 $t0$  & 116  & $4.46 \pm 3.05$  & $3.93 \pm 2.82$ & $11.83 \%$ \\ \hline
		us101 $t1$  & 93   & $5.13 \pm 2.92$  & $3.54 \pm 1.96$ & $31.03 \%$ \\ \hline
		us101 $t2$  & 133  & $5.11 \pm 3.15$  & $3.84 \pm 2.89$ & $24.94 \%$ \\ \hline
		highD    	& 55   & $3.40 \pm 2.38$  & $3.36 \pm 2.34$ & $1.18 \%$  \\ \hline
	\end{tabular}
    \vspace{-0.25cm}
\end{table}
\begin{table}[t]
	\footnotesize
	\centering
    \captionsetup{skip=0pt}
	\caption{\footnotesize Model Performance Comparison on Testing Datasets}
	\label{tab:results-combined-test}
	\begin{tabular}{|l|l|l|l|l|l|}
		\hline
        Highway  & \# Traj. & $\overline{MEE}_{w} \downarrow$ & $\overline{MEE}_{w+} \downarrow$ & \% Imp $\uparrow$    \\ \hline
		i80 $t0$	& 23  &	$3.21 \pm 2.47 $ & $3.08 \pm 2.46$ &  $4.11 \%$\\ \hline
		i80 $t1$ 	& 40  &	$4.43 \pm 2.78 $ & $4.38 \pm 2.83$ &  $1.05 \%$\\ \hline
		i80 $t2$	& 27  &	$3.63 \pm 2.21 $ & $3.54 \pm 2.12$ &  $2.48 \%$\\ \hline
		us101 $t0$  & 20  &	$5.50 \pm 2.82 $ & $4.85 \pm 2.13$ &  $11.78 \%$\\ \hline
		us101 $t1$  & 22  &	$6.12 \pm 4.93 $ & $5.27 \pm 4.06$ &  $13.92 \%$\\ \hline
		us101 $t2$  & 23  &	$3.81 \pm 1.56 $ & $3.51 \pm 2.63$ &  $8.03 \%$\\ \hline
		highD    	& 8   & $4.39 \pm 2.31 $ & $4.13 \pm 2.25$ &  $6.05 \%$\\ \hline
	\end{tabular}
\vspace{-0.5cm}
\end{table}
Tab.~\ref{tab:results-combined-train} summarizes trajectory similarity results on the training sets and Tab.~\ref{tab:results-combined-test} on the test sets in addition to the number of trajectories used for training and testing, respectively.
For each dataset, the third column summarizes the average and standard deviation of the MEE, in meters, between trajectories that are optimal with respect to the baseline reward and the corresponding expert trajectories in the dataset. Likewise, the fourth column summarizes the same values for trajectories optimal with respect to the unpredictability-aware reward.
The last column presents the improvement in average MEE of the unpredictability-aware model over the baseline, as a percentage of $\overline{MEE}_{w}$. On the test sets, we observe an MEE improvement between  1 to 13\% in the models incorporating the unpredictability feature and a weighted average improvement of $5.86\%$ across all test datasets.

\begin{figure}[t]
    \includegraphics[width=\columnwidth]{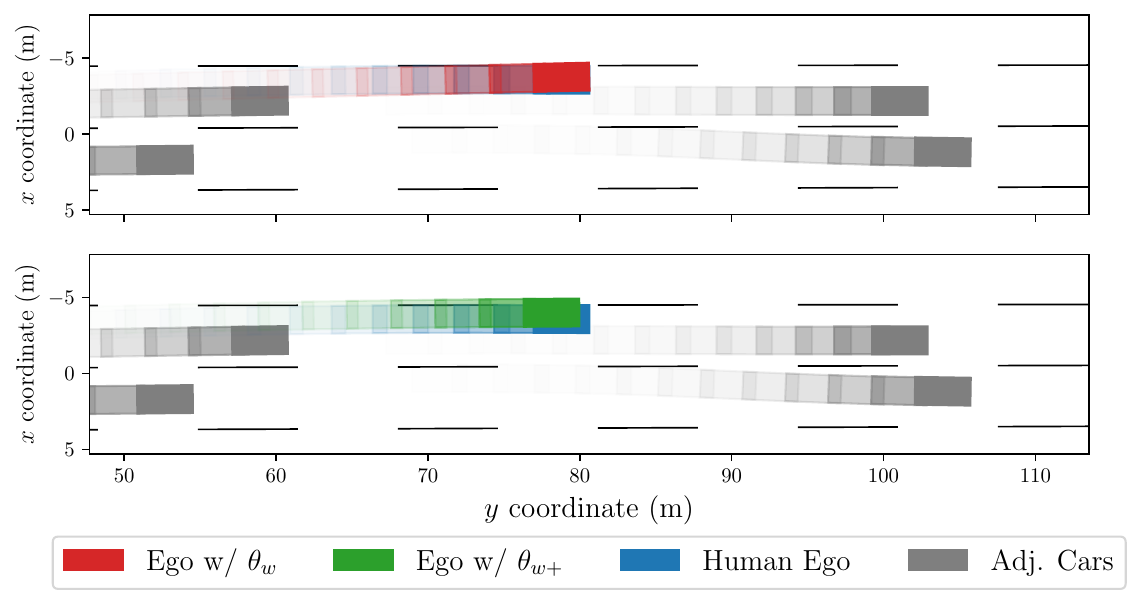}
    \centering
    \caption{Snapshot at $t=5.5s$ for baseline (red) unpredictability-aware (green) trajectories, compared to the human in the test set (blue). In this scenario none of the adjacent vehicles are unpredictable. The unpredictability-aware model (green) is very similar the baseline (red).}
    \label{fig:low_unpred_us101_t0_18}
    \vspace{-0.7cm}
\end{figure}
We observe that the trajectories generated using the reward function with unpredictability are consistently closer to the human expert trajectories compared to the baseline both on the training datasets (Tab.~\ref{tab:results-combined-train}) and on the test datasets (Tab.~\ref{tab:results-combined-test}). Note that the results are generated using diverse trajectory datasets from North America and Europe, and contain trajectories with variable levels of traffic congestion, capturing the diversity of human highway driving behavior.

In Fig.~\ref{fig:test_set_avg}, we plot the states of ego-car lane change trajectories from the humans in the US-101 $t1$ test set, as well as the trajectories generated using the baseline reward, $\mathbold{\theta}_{w}$, and the reward that incorporate unpredictability, $\mathbold{\theta}_{w+}$. For each state value, we plot the mean and 3-$\sigma$ bounds at each time step for all trajectories in the test dataset. We observe that for all state values, both models succeed in fitting the human trajectory data reasonably well, and both models succeed in traveling to the target lane. We note that incorporating unpredictability results in a better match with the lateral $x$ coordinate information indicating that in this dataset, the model with unpredictability is more hesitant to move into the target lane, similar to the human trajectory data. This result is consistent with the numerical improvement of $13.9\%$ listed in Tab.~\ref{tab:results-combined-test}. We believe the variation in improvement between the datasets can be attributed to the uneven distribution of unpredictable scenarios.
\begin{figure}[t]
    \includegraphics[width=\columnwidth]{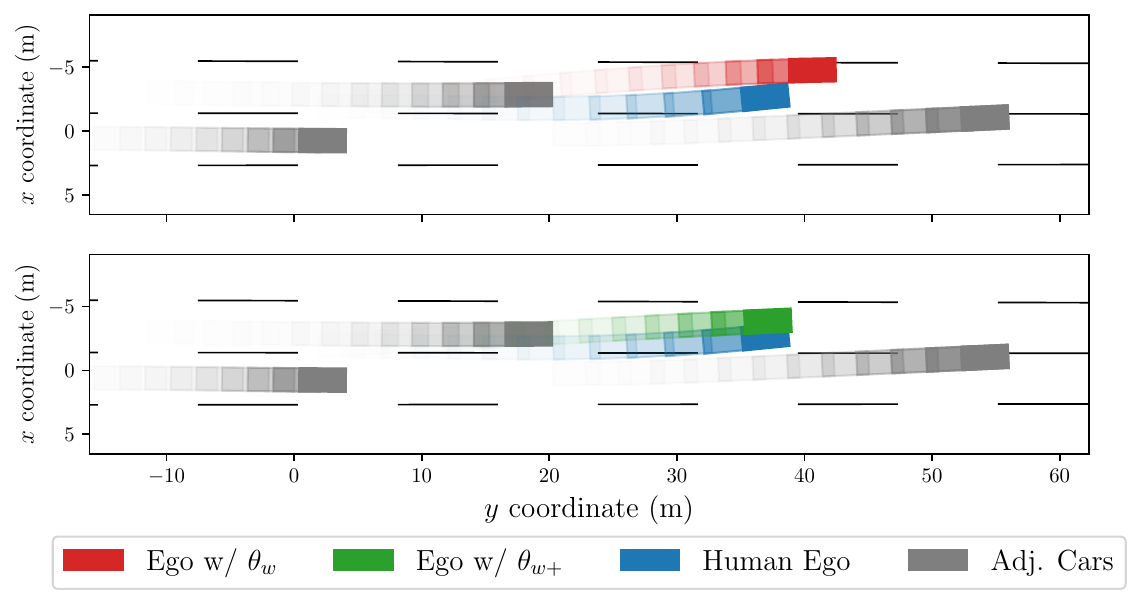}
    \centering
    \caption{Snapshot at $t=3.2s$. In this scenario, the adjacent vehicle preceding the ego car is behaving unpredictably by changing lanes in front of the ego car, who is also changing lanes. The unpredictability-aware model (green) delays changing into the target lane and maintains a higher distance from the preceding vehicle than the baseline (red), similar to the human (blue).}
    \label{fig:high_unpred_us101_t1_12}
    \vspace{-0.6cm}
\end{figure}
\vspace{-0.10cm}
\subsection{Qualitative Results}
\vspace{-0.05cm}
We also conduct a qualitative analysis of particular lane change scenarios from the test datasets with variations in the unpredictability of adjacent vehicles.%
\footnote{Animations of all scenarios can be found at \href{https://tiny.cc/unpredi}{\texttt{tiny.cc/unpredi}}.} %
As described in the motivating example in Fig.~\ref{fig:high_unpred_artificial}, we expect trajectories from the model incorporating unpredictability to keep a greater distance from an erratic or unpredictable adjacent vehicle compared to the baseline and a similar distance in the absence of unpredictability.
Fig.~\ref{fig:low_unpred_us101_t0_18} demonstrates a lane change scenario with low unpredictability for all the gray adjacent vehicles. We can observe that both models are very similar to the human trajectory, the green ($\mathbold{\theta}_{w+}$) and red ($\mathbold{\theta}_{w}$) vehicles are both very close to the blue (human).

Figs. \ref{fig:high_unpred_us101_t1_12}-\ref{fig:high_unpred_i80_t2_0} illustrate trajectories with an unpredictable adjacent car.
In Fig.~\ref{fig:high_unpred_us101_t1_12}, the preceding car is also moving to the target-lane simultaneously to the ego-car's lane change. We can see that the green vehicle ($\mathbold{\theta}_{w+}$) delays entering the target lane, similar to the blue human car, while the red vehicle ($\mathbold{\theta}_{w}$) does not. We make a similar observation even when the unpredictable adjacent vehicle is behind the ego car in the target lane (as in Fig.~\ref{fig:high_unpred_i80_t1_6}). In this scenario, the adjacent vehicle accelerates just as the human is changing into the target lane. While the green car ($\mathbold{\theta}_{w+}$) delays entering the target lane, similar to the blue human car, the red car ($\mathbold{\theta}_{w}$) quickly enters the target lane.
Like these two examples, on average, $\mathbold{\theta}_{w+}$ trajectories are more similar to human ones compared to $\mathbold{\theta}_{w}$. However, we also observed some scenarios where this is not the case for the whole trajectory. In Fig.~\ref{fig:high_unpred_i80_t2_0}, early in the lane change (\ref{fig:high_unpred_i80_t2_0_early}) we observe that the green car ($\mathbold{\theta}_{w+}$) delays entering the target, consistent with the other two scenarios. Here, however, the preceding car in the target lane behaves unpredictably by increasing its speed. %
The headway of the ego-car is increasing (\ref{fig:high_unpred_i80_t2_0_late}), yet the green vehicle ($\mathbold{\theta}_{w+}$) maintains its cautious distance: the increasing headway is counteracted by the unpredictability of the preceding car's acceleration.

In all the presented scenarios we observe that the trajectories generated using the unpredictability-aware behave more cautiously by staying farther away compared to the baseline. Note that on average across each test dataset, the unpredictability-aware trajectories are also more similar to the human data than the less-cautious baselines (recall Tab.~\ref{tab:results-combined-test}); the instances of over-cautiousness are a minority.

\subsection{Limitations and Future Work}
\begin{figure}[t]
    \includegraphics[width=\columnwidth]{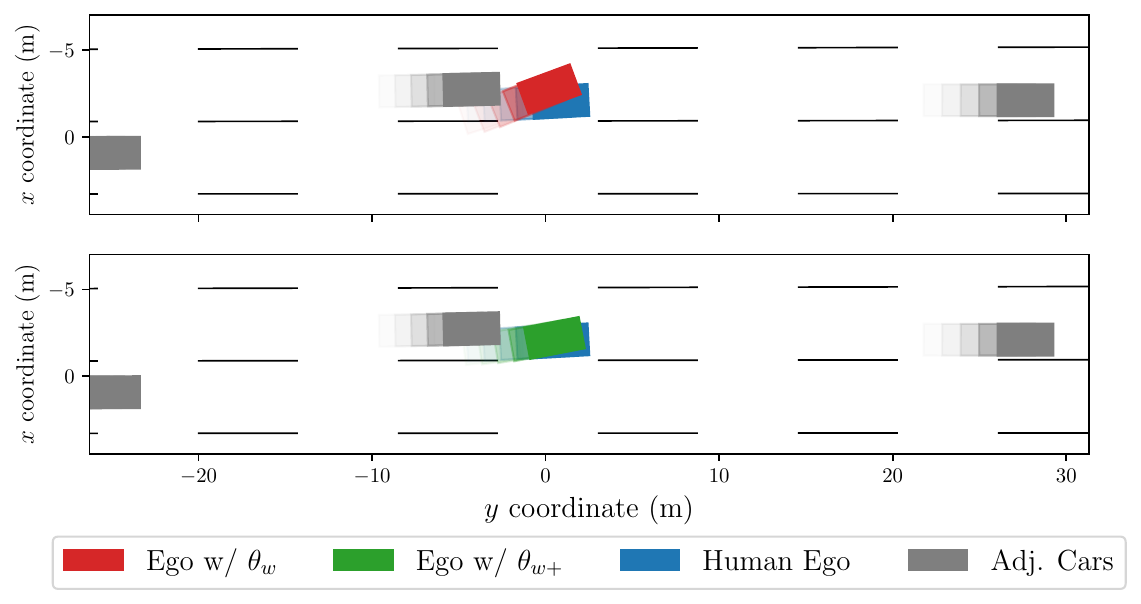}
    \centering
    \caption{
	Snapshot at $t=0.5s$. In this scenario, the adjacent vehicle in the target lane speeds up to not allow the ego vehicle in the target lane. The unpredictability-aware model (green) delays entering the target lane, similar to the human (blue), while the baseline model (red) cuts in front of the adjacent car.}
    \label{fig:high_unpred_i80_t1_6}
    \vspace{-0.5cm}
\end{figure}
\begin{figure}[t]
	\begin{subfigure}[]{\columnwidth}
		\includegraphics[width=\columnwidth]{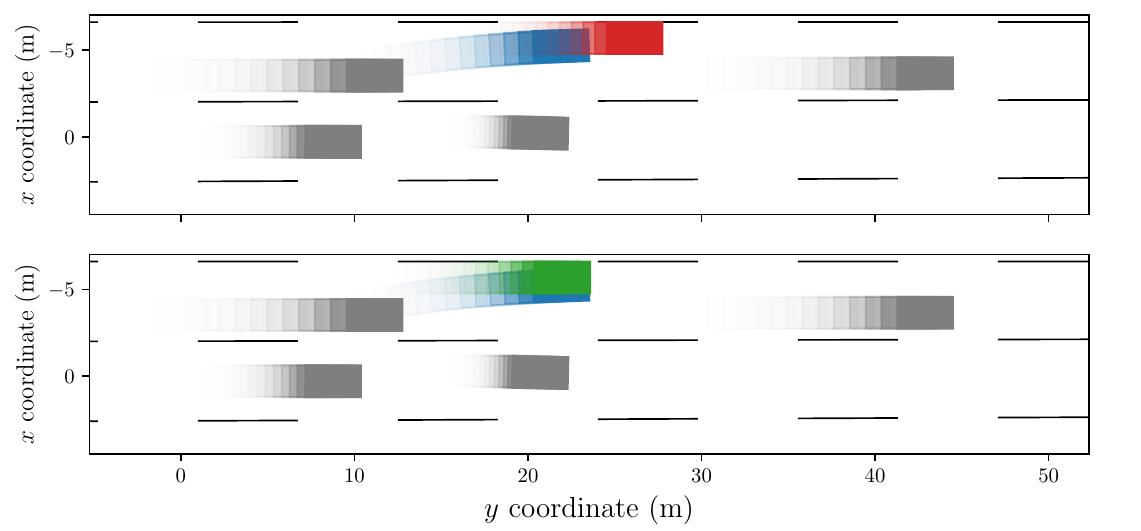}
		\captionsetup{skip=0pt}
		\caption{Snapshot at $t=3.5s$}
		\label{fig:high_unpred_i80_t2_0_early}
		\vspace{+0.2cm}
	\end{subfigure}
	\begin{subfigure}[]{\columnwidth}
		\includegraphics[width=\columnwidth]{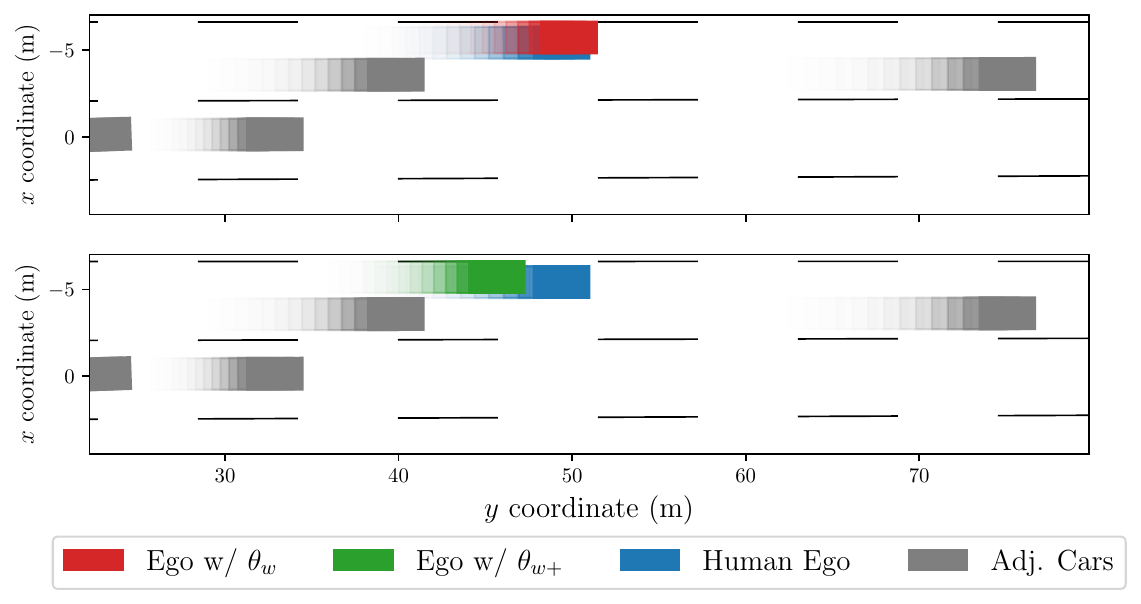}
		\captionsetup{skip=0pt}
		\caption{Snapshot at $t=7.0s$}
		\label{fig:high_unpred_i80_t2_0_late}
	\end{subfigure}
    \caption{
	Snapshots at $t=3.5s$ (a) and $t=7.0s$ (b) of the same scenario. The vehicle preceding the ego car behaves unpredictably by increasing its speed. Early in the trajectory (a) The unpredictability-aware model (green) delays entering the target lane, similar to the human car (blue) and unlike the baseline (red). Later in the trajectory (b), it continues to maintain a higher distance from the preceding vehicle despite the preceding car getting farther away, unlike the human (blue).}
	\label{fig:high_unpred_i80_t2_0}
    \vspace{-0.5cm}
\end{figure}
Although we can demonstrate an improvement in fitting the human data by using the unpredictability-aware reward, the scope of our study has limitations. %
First, our metric of unpredictability relies on the output of an off-the-shelf prediction model. %
Other prediction models or notions of unpredictability (e.g. covariance  \cite{fridovich2020confidence, Sun2021OnCE}) may have variable success in generating values that are effective in the reward that we present. %
An evaluation of such different notions is one direction for future work.

Second, our formulation structures reward functions as linear combinations of non-linear features, and the unpredictability-aware reward contains an extra feature versus the baseline. %
With this structure, one could argue that the additional degree of freedom bestowed by the extra feature means that \textit{any} feature that is sufficiently uncorrelated from those in the baseline could result in an improved fit. %
Although our quantitative results and qualitative analysis on simulated (Fig.~\ref{fig:high_unpred_artificial}) and real scenarios (Figs.~\ref{fig:low_unpred_us101_t0_18}-\ref{fig:high_unpred_i80_t2_0}) serve as an encouraging indication that unpredictability may be a cause, our study is limited in showing a definitive link. A further step is to also evaluate more diverse reward structures, for example, with a nonlinear combination of features \cite{SergeyLevineZoranPopovic2010} or neural net representation \cite{Wulfmeier2015}.

Third, we found that erratic or unpredictable behavior was rare in the highway lane changes that we used. A direction of future investigation is to extract driving scenarios where more unpredictable behavior can be found (e.g. from interactive driving datasets \cite{ettinger2021waymo}) and quantitatively analyze the effect of unpredictability in those particular occurrences, in addition to the average across all scenarios.
\section{Conclusion} \label{sec:conclusion}
In this paper, we proposed {\it unpredictability} as a feature in a reward function that can generate lane change trajectories. %
We used a performance metric of an off-the-shelf trajectory prediction model as a measure of the unpredictability of an adjacent driver, %
We analyzed human lane change behavior using an Inverse Reinforcement Learning (IRL) approach and show that incorporating this unpredictability measure can produce a better fit of human trajectories. %
We also provide qualitative insights into how the unpredictability feature may have an influence on the generated trajectories.
We believe our results encourage further investigation on the role of unpredictability in generating driving behavior.

\bibliographystyle{IEEEtran}
\bibliography{root}

\end{document}